\documentclass[10pt,twocolumn,letterpaper]{article}

\usepackage{cvpr}
\usepackage{times}
\usepackage{epsfig}
\usepackage{graphicx}
\usepackage{amsmath}
\usepackage{amssymb}
\usepackage{mmstyles}
\usepackage{bbm}
\usepackage{soul}
\usepackage{multirow}
\usepackage{array}
\usepackage{pifont}
\usepackage{authblk}
\usepackage{changes}

\newcolumntype{C}[1]{>{\centering\let\newline\\\arraybackslash\hspace{0pt}}m{#1}}

\graphicspath{{figures/}}
\DeclareGraphicsExtensions{.pdf}

\usepackage[pagebackref=true,breaklinks=true,letterpaper=true,colorlinks,bookmarks=false]{hyperref}

\cvprfinalcopy 

\def\cvprPaperID{1097} 

\begin{document}

\title{Self-Supervised Learning via Conditional Motion Propagation}

\author[1]{Xiaohang Zhan}
\author[1]{Xingang Pan}
\author[1]{Ziwei Liu}
\author[1]{Dahua Lin}
\author[2]{Chen Change Loy}
\affil[1]{CUHK - SenseTime Joint Lab, The Chinese University of Hong Kong}
\affil[2]{Nanyang Technological University}
\affil[1]{\tt\small \{zx017, px117, zwliu, dhlin\}@ie.cuhk.edu.hk}
\affil[2]{\tt\small ccloy@ntu.edu.sg}

\maketitle



\begin{abstract}

Intelligent agent naturally learns from motion. Various self-supervised algorithms have leveraged motion cues to learn effective visual representations. The hurdle here is that motion is both ambiguous and complex, rendering previous works either suffer from degraded learning efficacy, or resort to strong assumptions on object motions. In this work, we design a new learning-from-motion paradigm to bridge these gaps. Instead of explicitly modeling the motion probabilities, we design the pretext task as a conditional motion propagation problem. Given an input image and several sparse flow guidance vectors on it, our framework seeks to recover the full-image motion. Compared to other alternatives, our framework has several appealing properties: (1) Using sparse flow guidance during training resolves the inherent motion ambiguity, and thus easing feature learning. (2) Solving the pretext task of conditional motion propagation encourages the emergence of kinematically-sound representations that poss greater expressive power. Extensive experiments demonstrate that our framework learns structural and coherent features; and achieves state-of-the-art self-supervision performance on several downstream tasks including semantic segmentation, instance segmentation, and human parsing. Furthermore, our framework is successfully extended to several useful applications such as semi-automatic pixel-level annotation. Project page: \url{http://mmlab.ie.cuhk.edu.hk/projects/CMP/}.

\end{abstract}


\section{Introduction}
\begin{figure}[t]
	\centering
	\includegraphics[width=\linewidth]{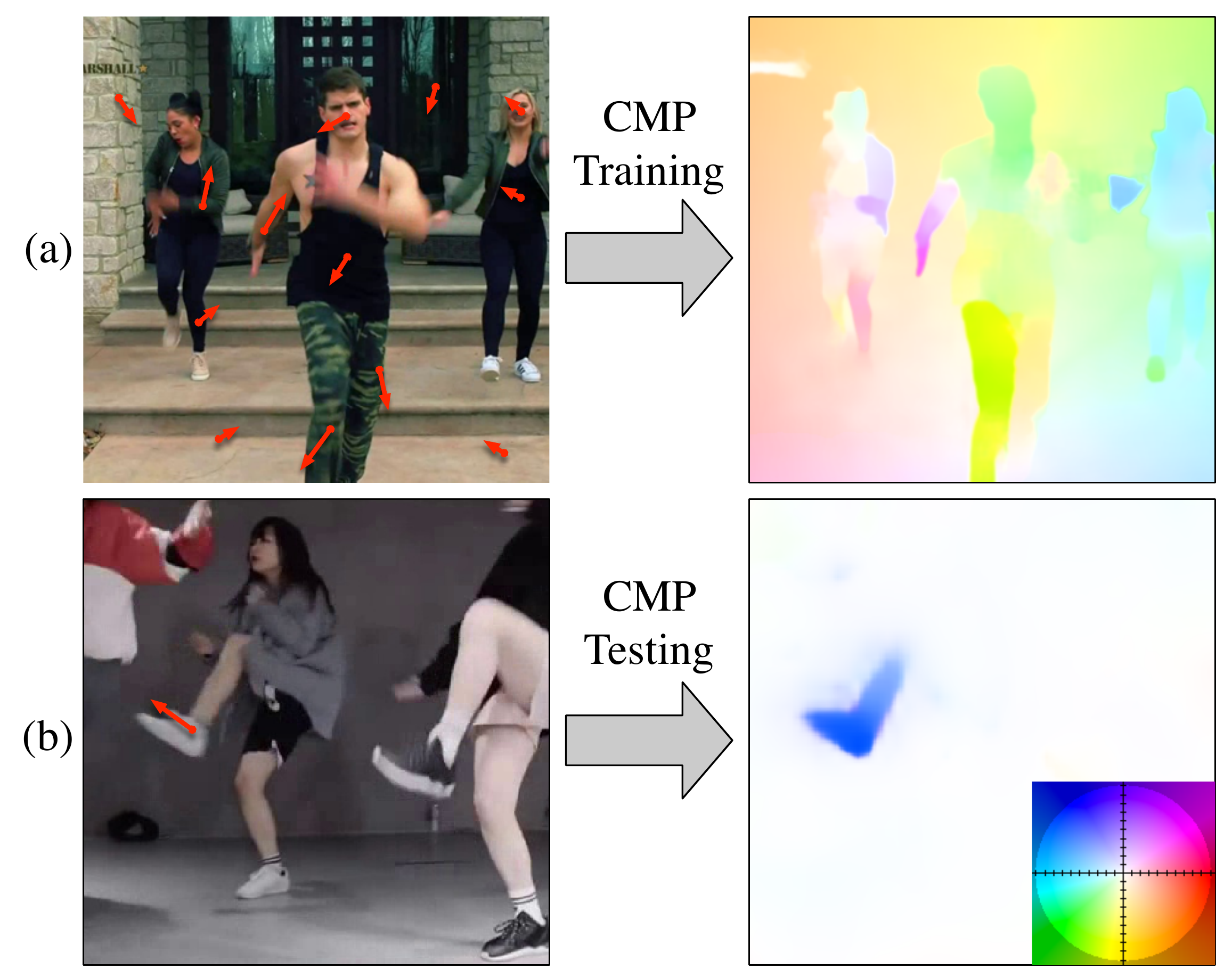}
	\caption{An illustration of our conditional motion propagation task. In training, the goal is to predict optical flow from a static image conditioned on sparse motion guidance. The guidance consists of sparse velocities sampled from the target optical flow with a ``watershed'' strategy (see Section~\ref{sec:guide}). In testing, the guidance can be arbitrary, and the model is able to predict kinematically-sound results. For example, as shown in (b), given a guidance on left foot, the model predicts that the shin is rotating. The optical flows are visualized with Middlebury color wheel, and should be viewed in \textbf{color}.}
	\label{fig:intro}
	\vspace{-10pt}
\end{figure}
Humans have a remarkable ability of gaining useful knowledge without direct supervision.
The visual world around us is highly structural, thus containing abundant natural supervisions to learn from.
In daily navigation, we constantly perform the task of visual prediction by hallucinating what's behind the corner.
The recently introduced self-supervised learning aims to empower machines with a similar capacity, learning without explicit annotations.
By carefully designing pretext tasks comprising natural supervisions, self-supervised learning learns effective representations that can be used for several downstream scenarios.

In comparison to static pretext tasks such as colorization~\cite{zhang2016colorful,larsson2017colorization} and inpainting~\cite{pathak2016context}, motion provides richer and more structural information for us to exploit.
The motion of a moving object generally indicates its kinematic properties, which further reveals its inner structure.
Previous works have leveraged the motion cues from two directions:
The first direction~\cite{walker2015dense,walker2016uncertain} is to learn image representations by predicting motion from static images.
For example, Walker~\etal~\cite{walker2015dense,walker2016uncertain} proposed to predict dense optical flow from a static image and use the learned features for action recognition.
%
%
However, since motion is inherently ambiguous, direct modeling of future motion creates large learning burden and sometimes results in unstable training.
The second direction~\cite{pathak2017learning,mahendran18cross} is to exploit the relationships between motion and objects to derive a motion-based constraining loss.
%
%
For example, Mahendran~\etal~\cite{mahendran18cross} assumed that pixels with similar features should have similar motions, and designed a cross pixel flow similarity loss to optimize the representations.
Though these methods have shown promising results, they made too strong assumptions on objects, \ie, all pixels on the same object should have similar motion.
%
%
%
%
However, most of the objects are intrinsically with high degrees of freedom.
For example, a person is an articulated object and a curtain is deformable.
We cannot claim that they still follow such simple assumption.

The ambiguity and complexity of motion pose great challenges on self-supervised algorithms.
In this work, to overcome these challenges and make better use of motion cues, we propose a new paradigm to leverage motion for representation learning.
The key idea is to define the pretext task as a \textit{Conditional Motion Propagation} (CMP) problem.
The framework is composed of an image encoder, a sparse motion encoder and a dense motion decoder.
As shown in Figure~\ref{fig:intro}, our task is to predict optical flow from a single image conditioned on sparse motion guidance.

Our approach has several merits.
Firstly, using sparse motion as guidance during training avoids the motion ambiguity problem, thus easing the pressure in representation learning.
Secondly, in order to recover dense optical flow from the given sparse motions, the image encoder must encode kinematically-sound properties so that the decoder is able to propagate motions from the guidance according to the properties.
Hence, in this way, the image encoder can automatically learn complex kinematic properties from motions, instead of predefining a specific relationship between motion and objects.
As shown in Figure~\ref{fig:intro} (b), in testing time, given an arbitrary guidance arrow, the CMP model produces kinematically reasonable results.
%
%
Leveraging such characteristics, CMP can also be applied to guided video generation and semi-automatic pixel-level annotation~\ref{sec:application}.

Thanks to the kinematically-sound representations learned by CMP, our method can benefit several downstream tasks, especially for segmentation tasks.
Our proposed CMP achieves state-of-the-art performance on several benchmarks under the condition of unsupervised pre-training, including PASCAL VOC 2012 semantic segmentation, COCO instance segmentation, and LIP human parsing.
We summarize our \textbf{contributions} as follows:
First, we propose a new paradigm to better leverage motion in representation learning and achieve promising performance on various benchmarks.
Second, our CMP model is capable of capturing kinematic properties of various objects without any manual annotations.
Third, the CMP model can be applied to guided video generation and semi-automatic annotation.

\begin{figure*}[htbp]
	\centering
	\includegraphics[width=0.9\linewidth]{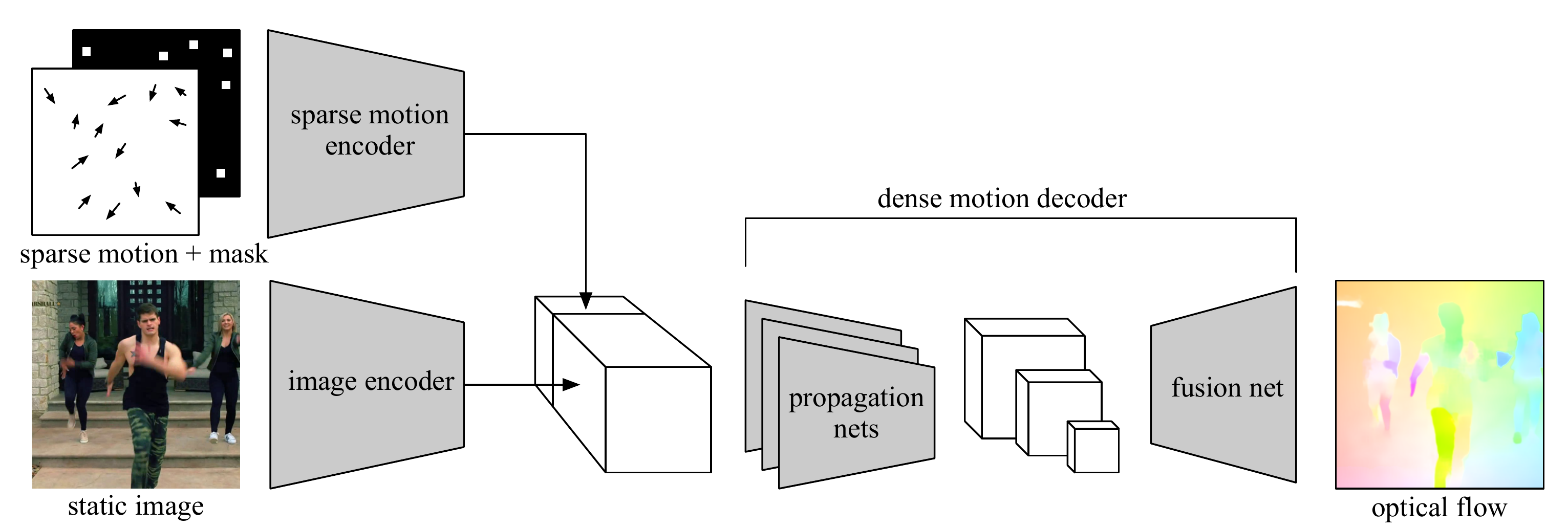}
	\caption{Our conditional motion propagation framework mainly contains three modules: sparse motion encoder, image encoder and dense motion decoder. Sparse motions are sampled from target optical flow with a ``watershed'' strategy illustrated in Section~\ref{sec:guide}. The target optical flow is extracted using off-the-shelf method.}
	\label{fig:framework}
	\vspace{-10pt}
\end{figure*}

\section{Related Work}
%
%
Self-supervised learning can be divided into two categories, respectively exploiting context and videos.

\noindent\textbf{Learning from Context.}
Context-based self-supervised learning methods typically distort or decompose the images and then learn to recover the missing information.
For instance, Doersch~\etal~\cite{doersch2015unsupervised} design a task to predict relative locations of patch pairs.
Pathak~\etal~\cite{pathak2016context} learn representations by image in-painting.
Noroozi~\etal~\cite{noroozi2016unsupervised} define jigsaw puzzles of image patches and train a CNN to solve them.
Zhang~\etal~\cite{zhang2016colorful} and Larsson~\etal~\cite{larsson2017colorization} learn features via colorizing gray images.
Gidaris~\etal~\cite{gidaris2018unsupervised} rotate images and then use CNN to predict the rotations.
%

%
\noindent\textbf{Learning from Temporal Consistency.}
For video-based representation learning, supervisions come from temporal information and thus images are usually undistorted.
Some of them rely on temporal consistency of contexts.
Mobahi~\etal~\cite{mobahi2009deep} make a temporal coherence assumption that successive frames tend to contain similar contents.
Jayaraman~\etal~\cite{jayaraman2016slow} train a CNN with a regularizer that feature changes over time should be smooth.
Wang~\etal~\cite{wang2015unsupervised} find corresponding pairs by visual tracking.
Other works~\cite{liu2017video,lee2017unsupervised,misra2016shuffle,wei2018learning} learn representations by synthesizing frames or predicting correct temporal order.

\noindent\textbf{Learning from Motion.}
Other video-based methods focus on motions to discover object-level information.
Pathak~\etal~\cite{pathak2017learning} use foreground segment masks extracted from videos as supervision.
Mahendran~\etal~\cite{mahendran18cross} assume that similar features should have similar motions, and design a cross pixel flow similarity loss to optimize the representation.
These works rely on a strong assumption, \ie, all pixels on the same object should have similar motion.
As mentioned before, most objects are intrinsically with high degrees of freedom.
Even the same object may have diverse motion patterns under different circumstances.
For example, pixels' motions on a bar are similar if it is shifting, but vary if it is rotating.

An alternative way to leverage motion for self-supervised learning is through performing optical flow prediction from static images.
Walker~\etal~\cite{walker2015dense} propose to predict dense optical flow from a static image.
And the follow-up work~\cite{walker2016uncertain} uses a Variational Auto Encoder to model the motion uncertainty.
However, due to the ambiguity of motion, it is a daunting task to predict motion without any hints, especially when coupled with camera ego-motion.
Recall that our target is to predict motion from static images conditioned on sparse motion guidance.
Hence motion forecasting is a degenerate case of our work when the amount of guidance points decreases to zero.
Using sparse motion as guidance during training avoids motion ambiguity problem, thus easing the difficulty in representation learning.


\section{Conditional Motion Propagation}
Our goal is to learn image representation by designing the pretext task as a conditional motion propagation problem.
Specifically, our training framework seeks to recover the full-image motion from static images conditioned on sparse motion guidance.

\subsection{Framework}

As shown in Figure~\ref{fig:framework}, the framework contains three modules: image encoder, sparse motion encoder, and dense motion decoder.

\noindent\textbf{Image Encoder.}
The image encoder is a standard backbone Convolutional Neural Network (CNN). After the CMP training completed, it serves as a pre-train model for the subsequent tasks.
CMP does not restrict the backbone architecture, though in our experiments the backbone is AlexNet or ResNet-50, depending on different target tasks.
We add an additional convolution layer at the top of the image encoder to encode the feature to 256 channels.

\noindent\textbf{Sparse Motion Encoder.}
It is a shallow CNN aiming at encoding the sparse motion into compact features.
It contains two stacked Conv-BN-ReLU-Pooling blocks and encodes sparse motion into 16 channels.
The spatial stride depends on the stride of the image encoder.
The inputs to the sparse motion encoder include: 1) The two-channels sparse optical flow as guidance sampled from the target optical flow using a ``watershed'' strategy discussed in Section~\ref{sec:guide}. The flow values of positions that are not sampled are set to zero. 2) A binary mask indicating the positions of selected guidance points. It serves to distinguish the sampled positions with zero motion and those unsampled positions.
We concatenate the sparse motion and the mask as a 3-channel input to the sparse motion encoder.
The motion and image features are concatenated and fed into the dense motion decoder.

\noindent\textbf{Dense Motion Decoder}.
The decoder is designed to propagate motion to the full image according to the encoded kinematic properties.
The decoder contains several propagation nets and a fusion net.
The propagation nets are CNNs with different spatial strides.
Those with larger spatial strides have larger receptive fields, hence they result in longer distances of propagation.
And those with smaller spatial strides focus on shorter distance, thus producing fine-grained results.
Each propagation net is composed of a max pooling layer with respective stride, and two stacked Conv-BN-ReLU blocks.
We design the propagation nets to be rather shallow, so as to force the image encoder to learn more meaningful information.
Finally, the output of propagation nets are up-sampled to the same spatial resolution and concatenated into the fusion net, a single convolution layer, to produce predictions.

\noindent\textbf{Loss Function.}
Optical flow prediction is typically regarded as a regression problem, as in ~\cite{dosovitskiy2015flownet}, since regression produces averagely accurate velocity values.
However, regression usually cannot produce discriminative gradients, and the results tend to be smoothed.
This issue could prevent us from learning good representations from scratch.
Fortunately, CMP does not need the output flow to be absolutely accurate.
Hence, we quantize the target flow and formulate it as a classification task.
Different from Walker~\etal~\cite{walker2015dense} who quantize optical flow by clustering, we adopt a simple yet efficient method.
We clip the target flow within a loose boundary, and partition the flow into $C$ bins linearly in $x$ and $y$ coordinates respectively.
They are then classified by two linear classifiers.
We use a cross-entropy loss separately for $x$ and $y$ flows. It is formulated as:
\begin{equation}
\begin{split}
&L_x = -\frac{1}{N}\sum_{i=1}^{N}\sum_{c=1}^{C}\left (\mathbbm{1}\left ( Q_i^x = c \right ) \log {P_i}_c^x \right ), \\
&L_y = -\frac{1}{N}\sum_{i=1}^{N}\sum_{c=1}^{C}\left (\mathbbm{1}\left ( Q_i^y = c \right ) \log {P_i}_c^y \right ),
\end{split}
\end{equation}
where $N$ is the total number of pixels, $P$ is the probability from SoftMax layer, $Q$ is the quantized labels, and $\mathbbm{1}$ is an indicator function.
We apply the same weight to $L_x$ and $L_y$.

\subsection{Guidance Selection}
\label{sec:guide}

\begin{figure}[t]
	\centering
	\includegraphics[width=0.9\linewidth]{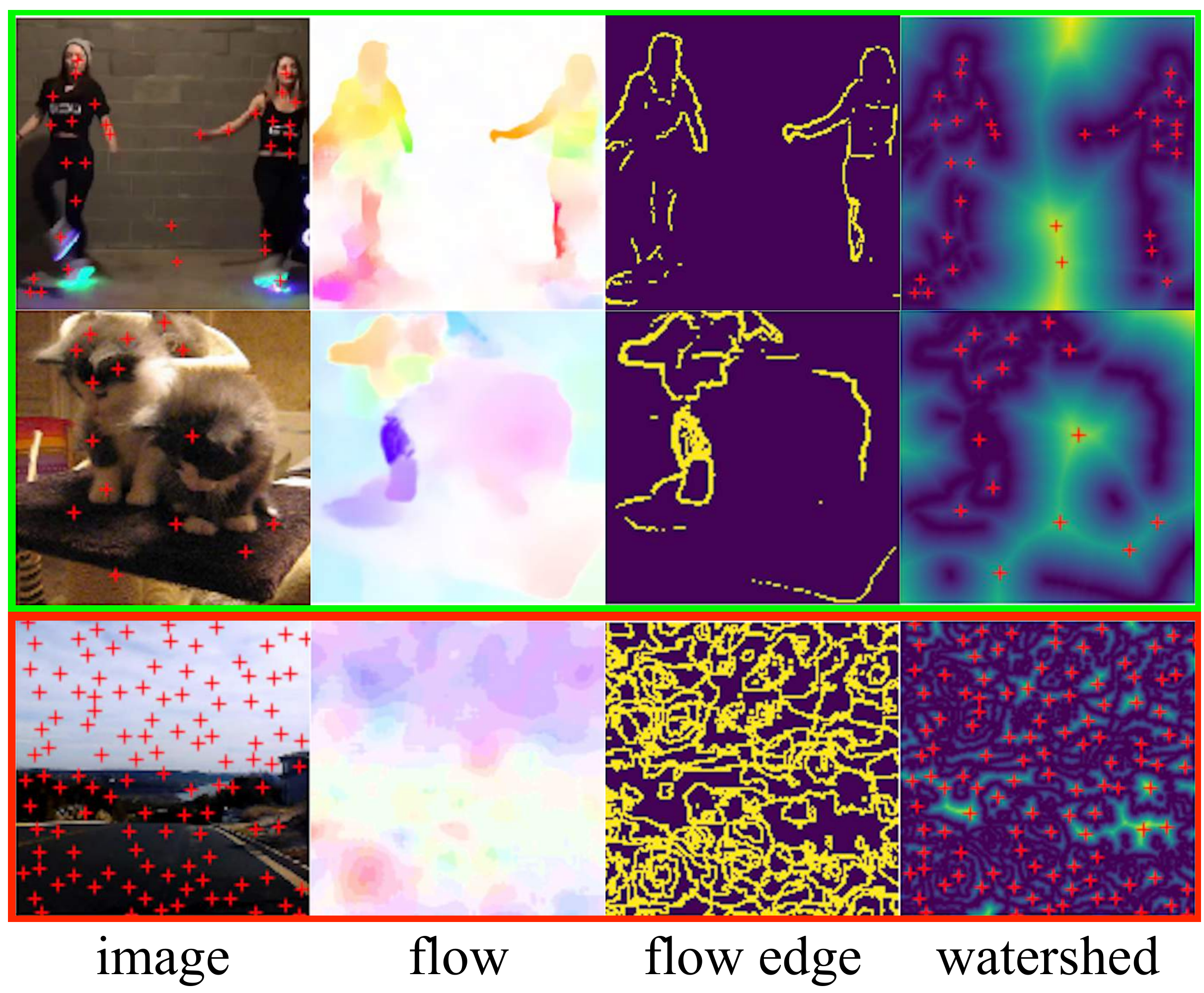}
	\caption{The figure shows how we sample guidance from optical flow. We first extract motion edges and then create a watershed map based on the edges. At last, we use NMS to obtain the key-points. Low-quality flow as shown in the third row results in a large number of key-points which instead eases the pressure to learn from those meaningless motions.}
	\label{fig:sampling}
	\vspace{-10pt}
\end{figure}

\noindent
\textbf{Sampling from Watershed.}
Sparse motion guidance is sampled from the target optical flow.
For effective propagation, those guidance vectors should be placed at some key-points where the motions are representative.
We adopt a watershed-based~\cite{beucher1979use} method to sample such key-points.
As shown in Figure~\ref{fig:sampling}, given the optical flow of an image, we first extract motion edges using a Sobel filter.
Then we assign each pixel a value to be the distance to its nearest edge, resulting in the topological-distance watershed map.
Finally, we apply Non-maximum Suppression (NMS)~\cite{canny1986computational} with kernel size $K$ on the watershed map to obtain the key-points.
We can adjust $K$ to control the average number of sampled points.
A larger $K$ results in sparser samples.
Points on image borders are removed.
With the watershed sampling strategy, all the key-points are roughly distributed on the moving objects. 
Since background motion actually reflects camera ego-motion, to avoid ambiguity in learning, we also add several grid points in each image.
The grid stride $G$ is used to adjust the density of grids.
For a good practice in our experiments, there are on average $13.5$ sampled guidance points in a $384\times384$ image.

\noindent
\textbf{Outlier Handling.}
In some cases, the optical flow may not be ideal, as shown in the third row of Figure~\ref{fig:sampling}.
The disordered flow edges create disconnected watersheds, which result in a large number of key-points selected.
However, it does not affect learning.
These image examples are actually easy cases, since the abundant guidance ease the pressure in learning those meaningless motions.
In other words, these examples with collapsed flows are ignored to some extent.
Hence, our framework is robust to the quality of optical flow.

%


\section{Experiments}

%
\noindent\textbf{Training Sets.}
CMP does not rely on a specific optical flow estimation method.
Considering that our datasets are million-level, we choose LiteFlowNet~\cite{hui2018liteflownet}, an extremely fast optical flow estimation tool to compute optical flows.
In this way, we prepare 4 training sets for CMP training.

\noindent
(a) \textbf{YFCC100m-Videos.}
YFCC100m contains about 700k in-the-wild videos.
We use the set of sampled frames provided by~\cite{pathak2017learning}, which originally contains 1.6M images from 205k video clips.
We use the image pairs with an interval of fewer than $10$ frames in sequence to compute optical flow.
For example, given a video clip containing 5 frames, and the frame IDs are $1$, $4$, $10$, $21$, $28$, we get $3$ image pairs, $\{1, 4\}$, $\{4, 10\}$, $\{21, 28\}$.
We use the first image in a pair and the computed flow to create an image-flow pair.
From those frames, we create about $1.26M$ image-flow pairs to form the training set (hereinafter referred to as ``YFCC'').

\noindent
(b) \textbf{YouTube9K.}
To show the benefits from more unlabeled data, we sample about 9,000 videos containing common objects from YouTube-8M~\cite{abu2016youtube}.
We sample the videos using keywords including ``bird'', ``cat'', ``dog'', \etc, which commonly exist in the visual world.
Since CMP is an unsupervised method, we do not use the tags in training.
In the same way, we create 1.96M image-flow pairs from these videos.

\noindent
(c) \textbf{VIP and MPII.}
Apart from the above datasets with general objects, we also use the videos in Video Instance-level Parsing (VIP) dataset~\cite{gong2017look}, and MPII Human Pose Dataset~\cite{andriluka14cvpr}.
They mainly contain multiple persons in various events.
The former results in 0.377M image-flow pairs and the latter 0.976M image-flow pairs.
We create the two datasets aiming at training a human-centric CMP model, so as to prove its effectiveness in understanding human kinematic properties.
Of course, we do not use any annotations from these two datasets.

\noindent\textbf{Training Details.}
We implement our framework with PyTorch~\cite{paszke2017automatic}.
We resize the image and flow so that the shorter side is $416$ and random crop to $384\times384$.
In guidance sampling, for YFCC and YouTube9K, we set the NMS kernel size $K$ to be $81$, and the grid stride $G$ to be $200$ pixels, which results in averagely $9.5$ watershed-based points and $4$ grid points per image.
For VIP and MPII, $K$ is $15$ and $G$ is $80$, when the images mainly contain multiple persons whose degrees of freedom are high.
We also analyze the influence of the number of guidance points in Sec.~\ref{sec:analysis}.

Training a CMP model is efficient.
For example, the ResNet-50 CMP model except that for human parsing is trained for $42K$ iterations, about $5.3$ epochs using YFCC.
It costs $7.5$ hours on 16 GTX-1080-Ti GPUs.
The AlexNet CMP model is trained for $140K$ iterations using YFCC+YouTube9K.
The CMP model for LIP human parsing is trained on all the 4 datasets for $70K$ iterations, about $2.1$ epochs.
The convergence is fast, hence we do not have to train CMP for an excessive number of epochs.
For all the cases, we use SGD with learning rate $0.1$, momentum $0.9$, weight decay $1e{-4}$.
We drop the learning by $10$ times at iteration $\frac{2}{3.5}I$ and $\frac{3}{3.5}I$, where $I$ is the total iteration.

\subsection{Evaluations for Representation Learning}
Using a CMP model as a pre-trained model, we show its effectiveness in feature learning by fine-tuning it on several downstream tasks covering semantic segmentation, instance segmentation, and human parsing.
Most of the previous works report their transfer learning results on AlexNet.
However, AlexNet is regarded as obsolete.
To build up-to-date self-supervised learning baselines, we also perform experiments with ResNet-50 in addition to AlexNet.
Hence, we adopt 4 benchmarks for evaluation, \ie, PASCAL VOC 2012 Semantic Segmentation (AlexNet), PASCAL VOC 2012 Semantic Segmentation (ResNet50), COCO 2017 Instance Segmentation (ResNet50), and LIP Human Parsing (ResNet50).
The fine-tuning details can be found in the supplementary materials.

\noindent\textbf{Baselines.}
For AlexNet, most previous works report their results on PASCAL VOC 2012 semantic segmentation.
However, previous studies do not support ResNet-50, hence we have to reimplement them.
For comparisons, we reimplemented recent works that use motion as supervision and have achieved impressive results.
Those methods include Pathak~\etal~\cite{pathak2017learning} and Walker~\etal~\cite{walker2015dense}.
Among them, Walker~\etal~\cite{walker2015dense} is a special case of CMP when the guidance points number is zero.
We optimize their hyper-parameters to achieve their best performances in these benchmarks.

\begin{table}[t]
	\centering
	\caption{PASCAL VOC 2012 benchmark for semantic segmentation, with AlexNet. Our method achieves state-of-the-art and surpasses the baselines by a large margin. Methods marked $^\dagger$ have not reported the results in their paper, hence we reimplemented them to obtain the results.}
	\begin{tabular}{ll|c}
		\hline
		\begin{tabular}{@{}l@{}}Method\\\small{(AlexNet)}\end{tabular} & Supervision & \begin{tabular}{@{}c@{}}VOC12 Seg.\\ \small{\% mIoU}\end{tabular} \\\hline\hline
		Krizhevsky~\etal~\cite{krizhevsky2012imagenet}  & ImageNet labels & 48.0 \\\hline
		Random       & -               & 19.8 \\
		Pathak~\etal~\cite{pathak2016context}       & In-painting     & 29.7 \\
		Zhang~\etal~\cite{zhang2016colorful}        & Colorization    & 35.6 \\
		Zhang~\etal~\cite{zhang2017split}       & Split-Brain     & 36.0 \\
		Noroozi~\etal~\cite{noroozi2017representation}  & Counting        & 36.6 \\
		Noroozi~\etal~\cite{noroozi2016unsupervised}    & Jigsaw          & 37.6 \\
		Noroozi~\etal~\cite{noroozi2018boosting}      & Jigsaw++        & 38.1 \\
		Jenni~\etal~\cite{jenni2018self}        & Spot-Artifacts  & 38.1 \\
		Larsson~\etal~\cite{larsson2017colorization}      & Colorization    & 38.4 \\
		Gidaris~\etal~\cite{gidaris2018unsupervised}      & Rotation        & 39.1 \\
		Pathak~\etal~\cite{pathak2017learning}$^\dagger$       & Video-Seg       & 39.7 \\
		Walker~\etal~\cite{walker2015dense}$^\dagger$ & Flow Prediction & 40.4 \\
		Mundhenk~\etal~\cite{mundhenk2018improvements}     & Context         & 40.6 \\
		Mahendran~\etal~\cite{mahendran18cross}    & Flow similarity & 41.4 \\\hline
		Ours & CMP 		   & \textbf{44.5} \\\hline
	\end{tabular}
	\label{tab:alexnet-seg}
\end{table}

\noindent\textbf{VOC2012 Semantic Segmentation (AlexNet)}.
Following previous works, we fine-tune the pre-trained weights on AlexNet for PASCAL VOC 2012 semantic segmentation task with FCN-32s~\cite{long2015fully} as the head.
As shown in Table~\ref{tab:alexnet-seg}, we achieve state-of-the-art performance with mIoU $44.5\%$ and surpass the baselines by a large margin.

\noindent\textbf{VOC2012 Semantic Segmentation (ResNet-50)}.
As shown in Table~\ref{tab:coco}, we achieve $59.0\%$ mIoU, with an $16.6\%$ improvement from a randomly initialized model.
The performance is also much higher than the baseline models.

\noindent\textbf{COCO Instance Segmentation (ResNet-50)}.
\begin{table}[t]
	\centering
	\caption{Results on PASCAL VOC 2012 Semantic Segmentaion validation set and COCO 2017 Instance Segmentation validation set, with ResNet-50.}
	\begin{tabular}{l|c|C{1.1cm}|c}
		\hline
		\multirow{2}{*}{\begin{tabular}[b]{@{}l@{}}Method\\\small{(ResNet-50)}\end{tabular}}
		& \multirow{2}{*}{\begin{tabular}{@{}c@{}}VOC12 Seg.\\ \small{\% mIoU}\end{tabular}}
		& \multicolumn{2}{c}{COCO17 \small(\% mAP)} \\\cline{3-4}
								 & & Det. & Seg. \\\hline\hline
		ImageNet~\cite{krizhevsky2012imagenet} 	& 69.0  & 37.2 & 34.1 \\\hline
		Random     								& 42.4 & 19.7 & 18.8	\\
		Pathak~\cite{pathak2017learning} 		& 54.6  & 27.7  & 25.8		\\
		Walker~\cite{walker2015dense}  			&	54.5	& 31.5 & 29.2 	\\\hline
		CMP (ours)     		& \textbf{59.0} & \textbf{32.3} & \textbf{29.8}  	\\\hline
	\end{tabular}
	\label{tab:coco}
\end{table}
We construct new baselines and the upper bound for self-supervised learning on COCO Instance Segmentation.
We use ResNet-50 as the backbone and Mask R-CNN~\cite{he2017mask} with FPN~\cite{lin2017feature} as the head.
As shown in Table~\ref{tab:coco}, we achieve $32.3\%$ bounding box mAP and $29.8\%$ mask mAP.
It indicates that CMP is an effective pre-training method for instance segmentation.

\noindent\textbf{LIP Human Parsing (ResNet-50)}.
Human parsing aims at partitioning a human image into pre-defined parts, \eg, head, arm, and leg.
Look-Into-Person (LIP)~\cite{gong2017look} is a large-scale benchmark for human parsing.
We perform comparisons on the validation sets of two sub-tasks, including LIP Single-Person Parsing and LIP Multi-Person Parsing.
As shown in Table~\ref{tab:lip}, we surpass baseline methods on both sub-tasks.
We further assemble our model with the model pre-trained on ImageNet, and observe higher performance than either of them.
It indicates that CMP pre-training is complementary with ImageNet pre-training.

\begin{table}[t]
	\centering
	\caption{LIP Human Parsing results on the validation set, with ResNet-50. The reported indicator is mIoU. The results marked with $^\star$ are obtained from the ensemble of our model and the model pre-trained on ImageNet.}
	\begin{tabular}{l|c|c}
		\hline
		\begin{tabular}{@{}l@{}}Method\\\small{(ResNet-50)}\end{tabular}& LIP-Single  & LIP-Multiple        \\\hline\hline
		ImageNet~\cite{krizhevsky2012imagenet}	& 42.5 & 55.4  	\\\hline
		Random     								& 32.5 & 35.0 	\\
		Pathak~\cite{pathak2017learning}  		& 36.6 & 50.9	\\
		Walker~\cite{walker2015dense}    		& 36.7 & 52.5 	\\\hline
		CMP (ours) 									& \textbf{40.2} & \textbf{52.9}  \\
		CMP$^\star$ (ours)  							& 42.9	&  55.8   \\\hline
	\end{tabular}
	\label{tab:lip}
\end{table}

\subsection{Further Analysis}
\label{sec:analysis}
\begin{figure}[t]
	\centering
	\includegraphics[width=\linewidth]{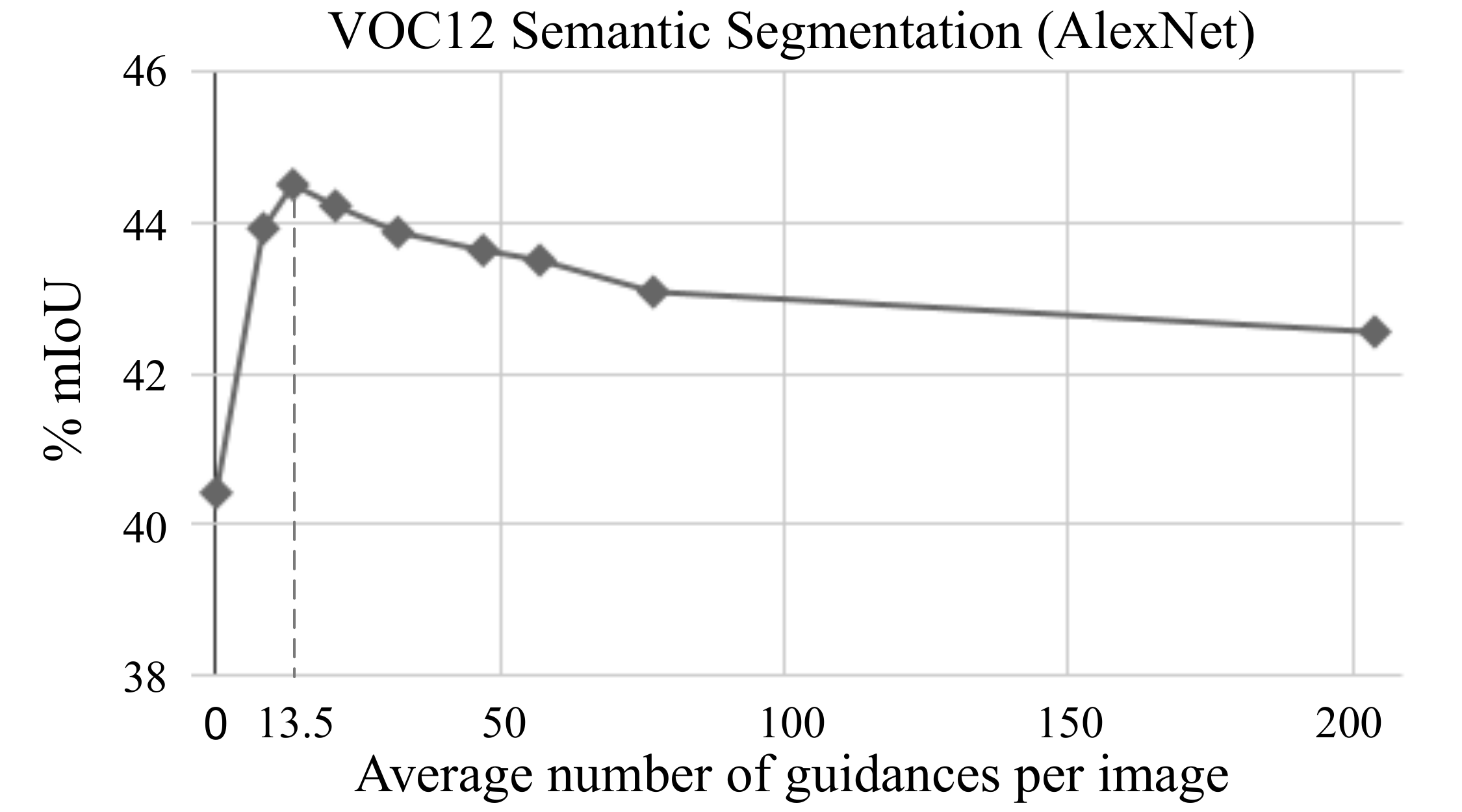}
	\caption{Influence of guidance number.}
	\label{fig:number}
\end{figure}

\noindent\textbf{Influence of Guidance Number.}
The number of guidance points is used to adjust the difficulty of pre-text CMP task.
An appropriate number of guidance points would allow a more effective CMP learning from images.
In this experiment, we adjust the NMS kernel size $K$ and grid stride $G$ to control the number of guidance points and perform an evaluation on the VOC 2012 semantic segmentation task with AlexNet.
As shown in Figure~\ref{fig:number}, the performance is low when the number of guidance is zero, and it is exactly the case in~\cite{walker2015dense}.
The peak occurs when the average number of guidance points is $13.5$.
As further guidance points join, the CMP task becomes easier.
Then the needed information to recover motions mostly comes from guidance rather than images.
Hence, the image encoder is weakened to capture essential information from images, and the performance drops.
Note that this optimal number of guidance points is related to the number of objects, and the degrees of freedom of each object in an image.
When the number of objects increases or the degrees of freedom goes higher, the number of guidance points should also increase accordingly.

\noindent\textbf{Influence of Propagation Nets.}
\begin{figure}[t]
	\centering
	\includegraphics[width=\linewidth]{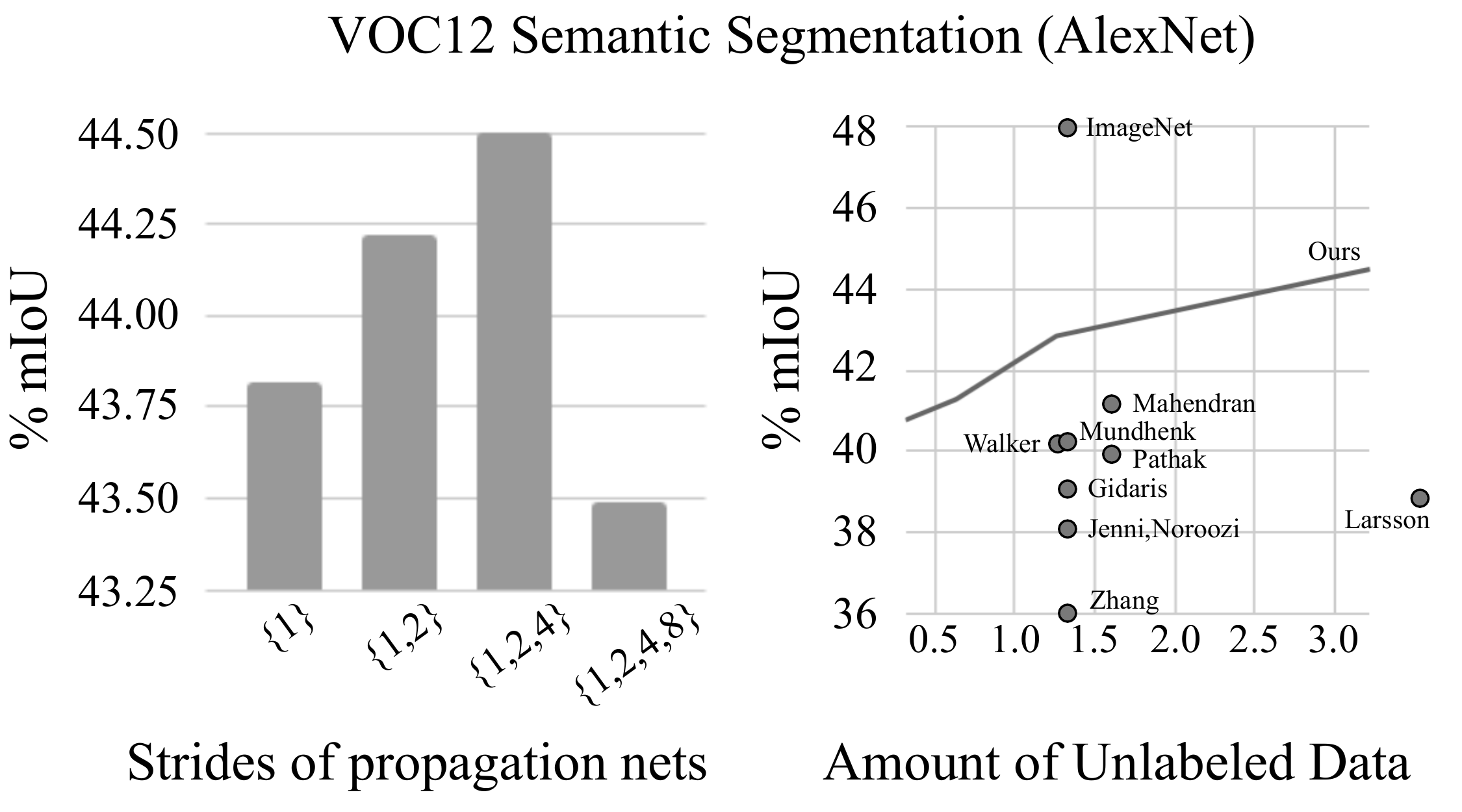}
	\caption{Influence of the combination of propagation nets and the amount of unlabeled data.}
	\label{fig:analysis}
	\vspace{-10pt}
\end{figure}
Recall that the propagation nets are the combination of several CNNs with different spatial strides.
We study the influence of different combinations of the propagation nets.
We implement $4$ propagation nets with spatial stride $1, 2, 4, 8$, and construct $4$ combinations, $\{1\}$, $\{1,2\}$, $\{1,2,4\}$, and $\{1,2,4,8\}$.
We test them on PASCAL VOC 2012 semantic segmentation with AlexNet.
As shown in Figure~\ref{fig:analysis}, an optimal combination occurs at $\{1,2,4\}$.
It indicates that the propagation nets with different strides form a collaborative group to solve the CMP problem effectively.
However, additional propagation net with overly large stride leads to the loss of spatial information while increasing the parameter count in the decoder, hence a combination of $\{1,2,4,8\}$ strides is worse.
Nevertheless, the performance is still much better than the baseline methods.

\noindent\textbf{Influence of the Amount of Unlabeled Data.}
We create $5$ training sets using $1/4, 1/2$ YFCC, full YFCC, and YFCC+YouTube9K.
The amount of data ranges from $0.32M$ to $3.22M$.
We test the AlexNet models trained respectively on these training sets on PASCAL VOC 2012 semantic segmentation task.
As shown in Figure~\ref{fig:analysis}, as the amount of unlabeled data increases, CMP achieves steady improvements.
The performance is much better than the baselines in a comparable amount of unlabeled data.

\begin{figure}[t]
	\centering
	\includegraphics[width=\linewidth]{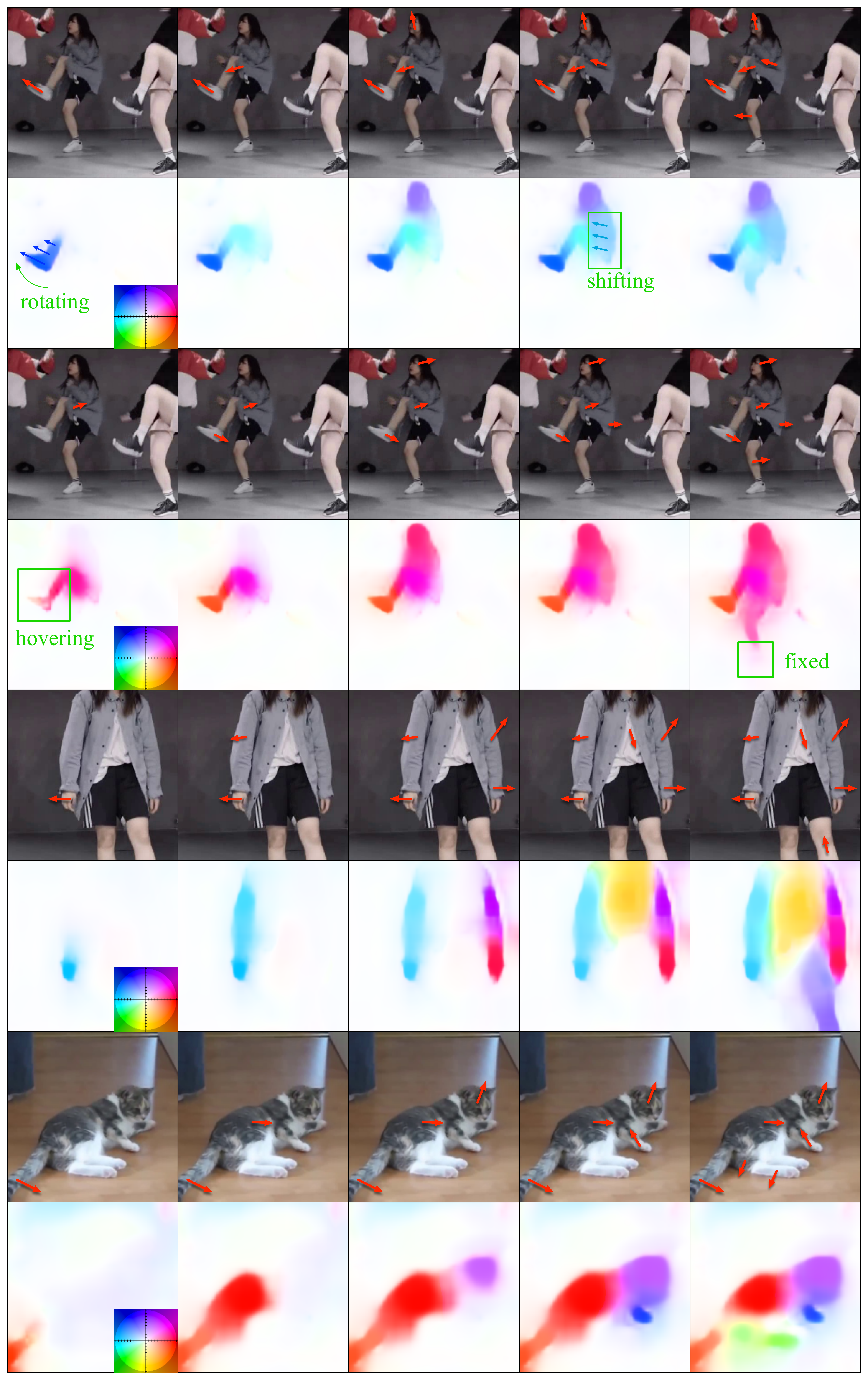}
	\caption{CMP testing results. In each group, the first row includes the original image and the guidance arrows given by users, the second row shows the predicted motion. The results demonstrate three characteristics of CMP: 1. CMP propagates motion on the whole rigid part. 2. CMP can infer whether a part is shifting or rotating (motion uniform if shifting, fading if rotating) as shown in the first group. 3. The results are physically feasible. For example, in the second group, given a single guidance vector on the left thigh, there are also responses on left shank and foot. It is due to the observation that the left leg is hovering. However, in the last column, although given a guidance vector on the right leg, the right foot keeps still because it is on the ground.}
	\label{fig:vlz}
	\vspace{-10pt}
\end{figure}

\noindent\textbf{CMP's Characteristics.}
Given a test image exclusive to the training set, we test a trained CMP model by giving arbitrary guidance vectors.
As shown in Figure~\ref{fig:vlz}, given an increasing number of guidance vectors, CMP infers more complete motions accordingly.
The results clearly reflect the structures of objects even with high degrees of freedom.
From the results, we observe three interesting characteristics of CMP:

\noindent 1) Rigidity-aware. Given a single guidance vector on a rigid part, \eg, head, forearm, or thigh, CMP propagates motion on the whole part.

\noindent 2) Kinematically-coherent. Given a guidance vector on a part, CMP is capable of inferring whether the part should be shifting or rotating. As shown in the first group in Figure~\ref{fig:vlz}, the body should be shifting, then it predicts uniform motion on the body, and the left leg should be rotating, hence the motion is fading.

\noindent 3) Physically-feasible. For example, in the first column of the second group in Figure~\ref{fig:vlz}, given a single guidance vector on the left thigh, there are responses on left thigh, shank, and foot. It is due to the observation that the left leg is hovering. However, in the last column, given a guidance vector on the right leg, the right foot keeps still, because it is on the ground.

Motion, though coarse and noisy, is the manifestation of kinematics and physics.
To achieve sensible motion propagation in complicated environments, our model must learn to imagine the intrinsic kinematic properties and physically-sound laws from static images.
It accounts for these three characteristics.
%

\subsection{Applications}\label{sec:application}

\begin{figure*}[t]
	\centering
	\includegraphics[width=\linewidth]{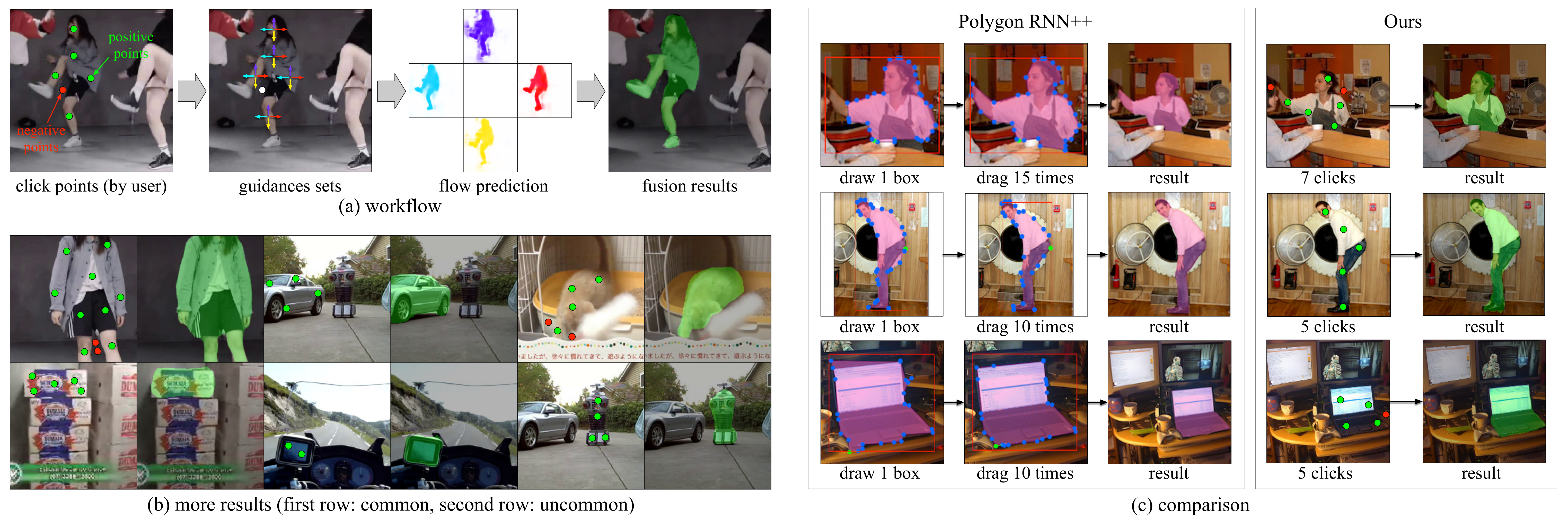}
	\caption{CMP for semi-automatic pixel-level annotation. (a) shows its workflow, where a user only needs to click several positive points (green) on the object, then the mask is automatically generated. If the mask covers some wrong areas, then the user clicks negative points (red) on the wrong areas, and the mask gets refined. (b) shows a single CMP model is able to assist users to annotate objects in any category, even the categories that the model has never seen. We compare our method with Polygon RNN++ in (c). For a fair comparison, we use the images from the web demo of Polygon RNN++. It requires a user to draw a bounding box at first and then drag the generated vertexes to refine. In some cases, it fails to capture the target object (second row). While our method does not require tedious dragging. It generates robust masks with only a few clicks.}
	\label{fig:annot}
\end{figure*}

\begin{figure}[t]
	\centering
	\includegraphics[width=\linewidth]{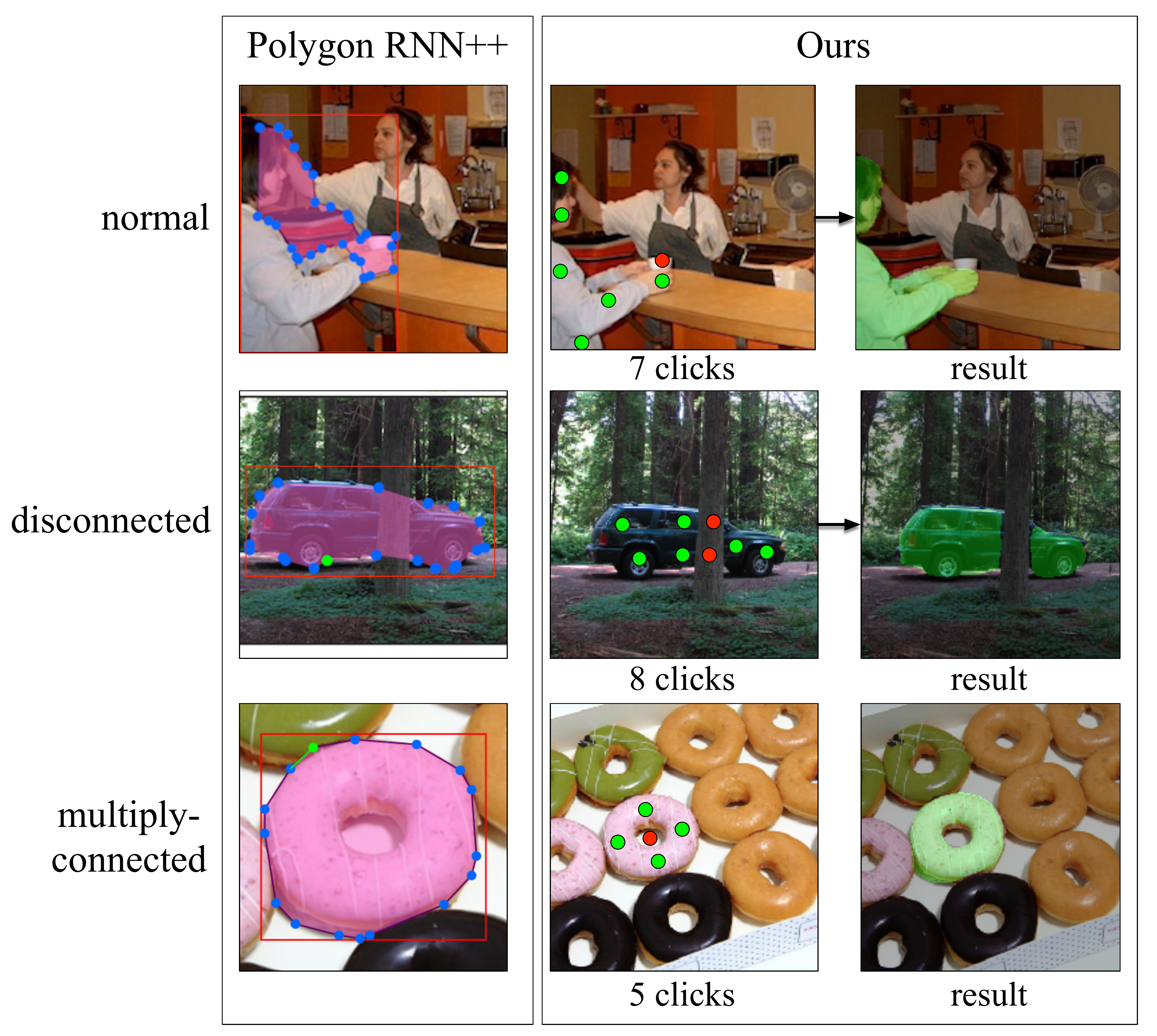}
	\caption{This figure illustrates the limitations of Polygon RNN++, and how CMP solves those cases.}
	\label{fig:fail}
\end{figure}

CMP shows its effectiveness in capturing structural kinematic properties of objects.
With such characteristics, several applications can be extended from a trained CMP model.
%
%
%
The image encoders for these applications are implemented with ResNet-50.

\noindent\textbf{CMP for Guided Video Generation.}
An interesting application of CMP is guided video generation.
With CMP, this application is reminiscent of marionette control.
Given an image and the guidance arrows from a user, we first use a CMP model to predict the optical flow and then warp the original image to produce the future frame.
In this way, we can create a sequence of frames by giving continuous guidance.
Since CMP is strong in perceiving rigid parts of an object from a single image, we can perform sophisticated marionette control on the image.
A demo video can be found in the project page~\footnote{Project page: http://mmlab.ie.cuhk.edu.hk/projects/CMP/}.

\noindent\textbf{CMP for Semi-automatic Pixel-level Annotation.}
We show that CMP can also assist pixel-level annotation.
Figure~\ref{fig:annot} (a) shows its workflow.
A user only needs to click several positive points on the object.
We make dummy guidance vectors on these points in different directions, then a CMP model predicts the optical flow in each direction.
Finally, we fuse the results to obtain the mask.
If the mask covers some wrong areas, then the user clicks negative points on the wrong areas.
For CMP, the negative points serve as the static guidance points with zero motion.
Hence, there will be no response around those negative points.
In this way, the mask gets refined.
Such interactive annotating mode allows the user to freely refine the mask via adding or deleting the two types of points.

Since CMP is an unsupervised method, it does not pre-define a specific category set like other semi-automatic annotation tools.
Instead, it captures the spatial structures of objects.
Hence, we can use it to annotate any unseen or uncommon objects, \eg, carton, rearview mirror, and robot, as shown in the second row of Figure~\ref{fig:annot} (b).

We compare our method with a state-of-the-art supervised semi-automatic annotating method, Polygon RNN++~\cite{acuna2018efficient}.
For a fair comparison, we test on the images from the web demo of Polygon RNN++.
As shown in Figure~\ref{fig:annot} (c), Polygon RNN++ requires a user to draw a bounding box at first, and then generates vertexes to form an initial mask.
However, the initial mask is usually imperfect.
The user needs to drag the vertexes to refine the mask.
In comparison, our method generates robust masks with only a few clicks.
The refinement is also simple and intuitive, conducted via interactive point-and-click to add or delete points.

In addition, Polygon RNN++ has some limitations as shown in Figure~\ref{fig:fail}:
1) In some cases, it fails to capture the target object.
2) It cannot correctly segment objects with disconnected regions (\eg, a car behind a tree.)
3) It cannot handle multiply-connected objects (\eg, a doughnut).
While our method can handle all of those cases by clicking positive and negative points.
The comparisons are summarized in Table~\ref{tab:annot}.
Note that Polygon RNN++ relies on supervised models, while our method is unsupervised without any manual annotations.

\begin{table}[t]
	\centering
	\caption{Comparisons with Polygon RNN++. ``sup'' and ``unsup'' stand for ``supervised'' and ``unsupervised''. ``MC'' and ``DC'' stand for whether they support ``multiply-connected'' and ``disconnected'' objects. Time per instance is tested on a randomly chosen subset from COCO dataset.}
	\begin{tabular}{l|c|c|c|c|c}
		\hline
		Method & model  & speed  & fail & MC & DC    \\\hline
		Polygon~\cite{acuna2018efficient} & sup	& 17.6s & 25/170 & \ding{56}   & \ding{56}	\\
		Ours  	& unsup & 10.2s & 0/170	& \ding{52} & \ding{52}	   \\\hline
	\end{tabular}
	\label{tab:annot}
\end{table}


\section{Conclusion}

To summarize, we propose a new self-supervised learning paradigm, Conditional Motion Propagation (CMP).
It learns effective visual representations for structural prediction. We achieve state-of-the-art performance in standard self-supervised representation learning benchmarks.
We also establish new benchmarks with ResNet-50 beyond just AlexNet.
CMP shows appealing characteristics in capturing kinematic properties of various objects with unlabeled data. We observe kinematically sound results when testing a CMP model.
Furthermore, CMP can be extended to several useful applications. For semi-automatic pixel-level annotation, we achieve encouraging usability when compared with a state-of-the-art supervised method.

\noindent
\textbf{Acknowledgement}: This work is partially supported by the Collaborative Research grant from SenseTime Group (CUHK Agreement No. TS1712093), the Early Career Scheme (ECS) of Hong Kong (No. 24204215), the General Research Fund (GRF) of Hong Kong (No. 14241716, 14224316. 14209217, 14236516, 14203518), and Singapore MOE AcRF Tier 1 (M4012082.020).

\clearpage
{\small
\bibliographystyle{ieee_fullname}
\bibliography{egbib}
}

\appendix
\section{Network Configurations}
\begin{figure}[!t]
	\centering
	\includegraphics[width=\linewidth]{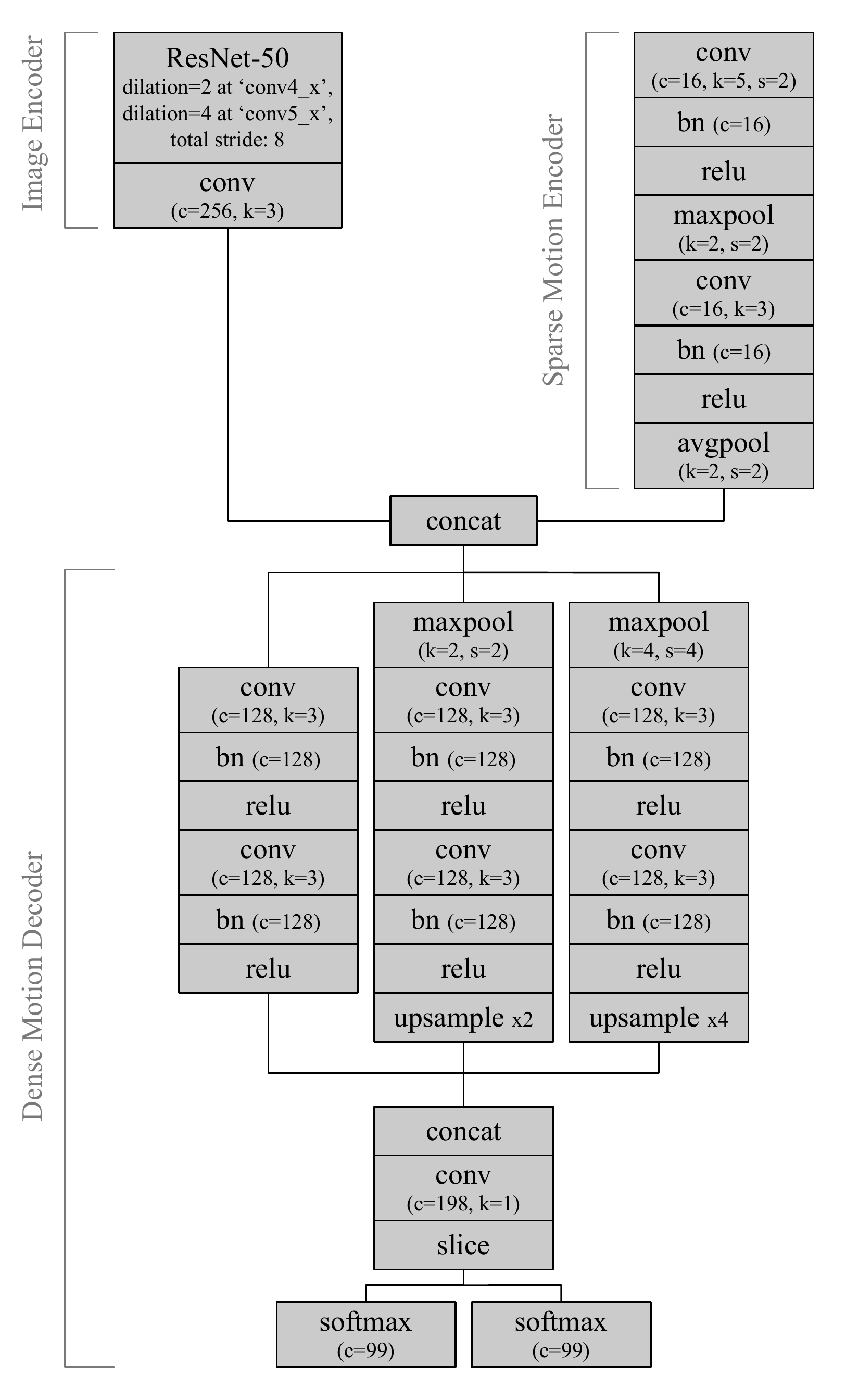}
	\caption{Network configurations, taking ResNet-50 for example. Notations ``conv4\_x'' and ``conv5\_x'' are the last two residual groups defined in ~\cite{he2016deep}. Parameters $c$, $k$ and $s$ stand for the number of output channels, kernel size and stride.}
	\label{fig:config}
\end{figure}

Here we illustrate the detailed configurations of the network, taking ResNet-50 for example.
As shown in Figure~\ref{fig:config}, the image encoder contains the backbone network and an additional convolution layer to encode images into features with $256$ channels.
To keep spatial structures in embedded features, we set the total stride in ResNet-50 to $8$, and dilations to $2$ and $4$ for ``conv4\_x'' and ``conv5\_x'', the last two residual groups defined in ~\cite{he2016deep}.

\section{Fine-tuning Details}
\noindent\textbf{VOC2012 Semantic Segmentation (AlexNet)}.
Following previous works, we fine-tune the pre-trained weights on AlexNet for PASCAL VOC 2012 semantic segmentation task with FCN-32s~\cite{long2015fully} as the head.
%
%
%
We remove the additional convolution layer of our image encoder, and fine-tune all the layers.
The initial learning rate is $0.01$ and it is decayed by $10$ times at $30K$, $48K$, $60K$ iterations. The total iteration is $66K$.

\noindent\textbf{VOC2012 Semantic Segmentation (ResNet-50)}.
We fine-tune the ResNet-50 CMP model for $33K$ iterations with an initial learning rate of $0.01$, with the polynomial learning rate decay strategy (power: $0.9$).
All the experiments including baselines, upper bound and our method use the same hyper-parameters.

\noindent\textbf{COCO Instance Segmentation (ResNet-50)}.
We construct new baselines and the upper bound for self-supervised learning on COCO Instance Segmentation.
We use ResNet-50 as the backbone and Mask R-CNN~\cite{he2017mask} with FPN~\cite{lin2017feature} as the head.
We use the same hyper-parameters across all the experiments, including an initial learning rate of $0.02$, learning rate decaying by $10$ times at epoch $10$ and $15$, and the total epoch is $16$.
Those hyper-parameters are expected to be fixed for future self-supervised learning studies.

\noindent\textbf{LIP Human Parsing (ResNet-50)}.
We perform a comparison on the validation sets of two sub-tasks, including LIP Single-Person Parsing and LIP Multi-Person Parsing.
The fine-tuning epochs are respectively $50$ and $120$ for these two tasks.
The initial learning rate is $0.01$, and the learning rate decay strategy is polynomial (power: $0.9$).
The hyper-parameters are kept the same across all the experiments.

\section{Evaluation on Detection}
We additionally perform experiments with VGG-16 to compare with a recent multi-task based self-supervised learning method~\cite{wang2017transitive} which achieves state-of-the-art with VGG on PASCAL VOC 2007 detection task.
We use the released pre-trained model of Wang~\etal~\cite{wang2017transitive} for detection and segmentation evaluation.
The evaluating experiments are conducted in the same circumstances.
As shown in Table~\ref{tab:vgg}, CMP does better in segmentation tasks than detection tasks, since CMP focuses on learning spatial structural representations.
\begin{table}[htbp]
	\centering
	\caption{Evaluation on VOC 2007 detection and VOC 2012 segmentation. Comparison with Wang~\etal~\cite{wang2017transitive}. For detection of Wang~\etal~\cite{wang2017transitive}, 63.2\% is reported and 57.0\% is reproduced.}
	\begin{tabular}{l|c|c}
		\hline
		& Det. (mAP) & Seg. (mIoU) \\ \hline
		ImageNet~\cite{krizhevsky2012imagenet}  & 67.3    & 64.1   \\ \hline
		Random   & 39.7    & 35.0    \\
		Wang~\etal~\cite{wang2017transitive} & 63.2 (57.0)  &  54.0   \\\hline
		CMP       &  56.8   &  57.6   \\ \hline
	\end{tabular}
	\label{tab:vgg}
\end{table}

\begin{figure}[t]
	\centering
	\includegraphics[width=\linewidth]{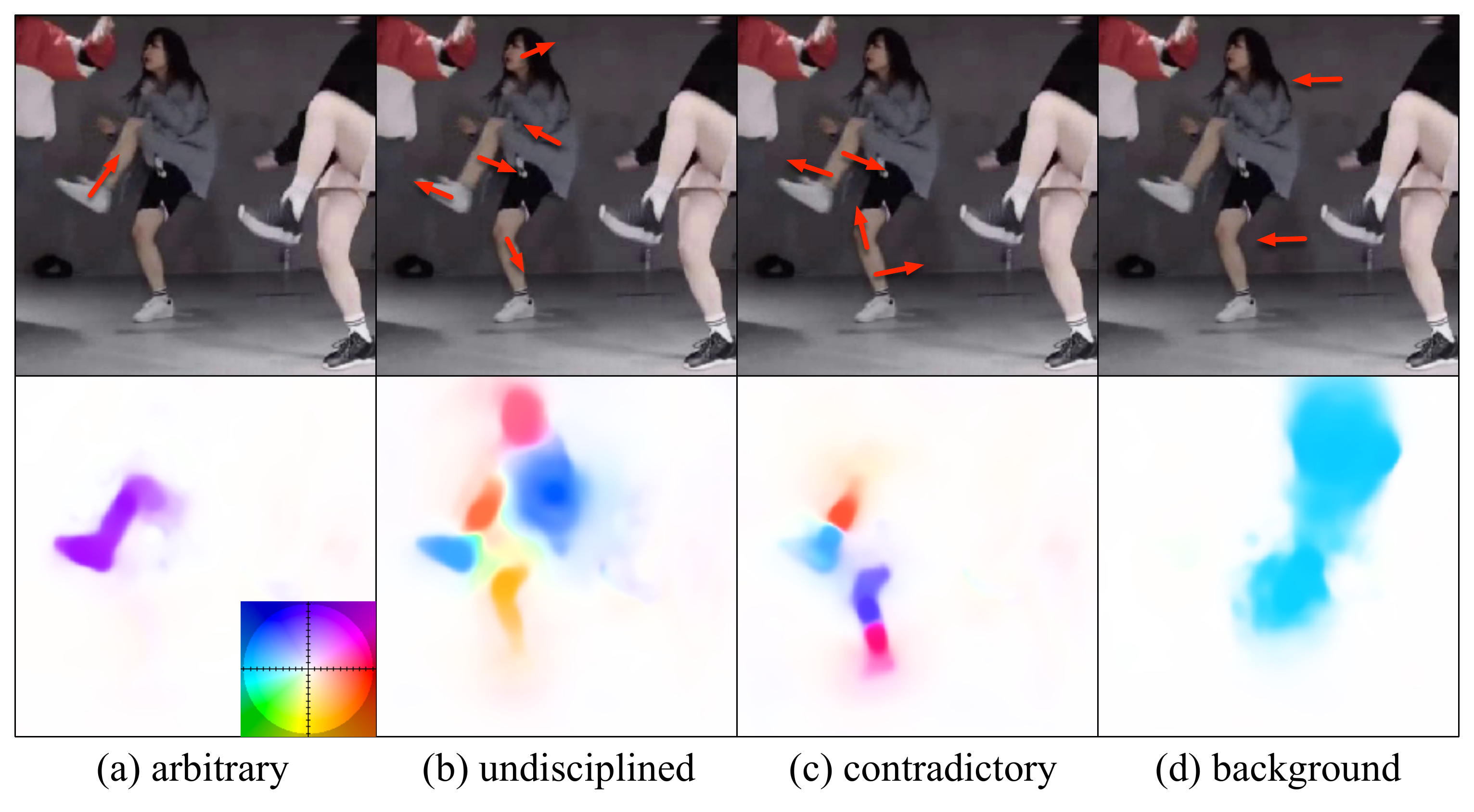}
	\caption{Noisy motion guidance.}
	\label{fig:noisy}
\end{figure}

\section{Visualizations}
\noindent\textbf{Testing with Noisy Guidance.}
To better understand CMP's ability of learning kinematic properties.
We deliberately give noisy guidance in testing.
As shown in Figure~\ref{fig:noisy},
(a) Given arbitrary guidance on a single point, rigidity awareness and physical feasibility still hold.
(b) Given a group of undisciplined guidance vectors, \ie, given random guidance vectors on different parts, these characteristics hold locally. The global kinematic coherent does not hold expectably, because the CMP model faithfully follows the given guidance, rather than over-fits the image to produce a plausible result.
(c) Given contradictory guidance, \ie, given two guidance vectors in different directions on a rigid part, the rigidity awareness does not hold anymore.
(d) Given outlying guidance on background, the motions are propagated within the background, while the foreground objects' optical flows are not affected.

\begin{figure}[h]
	\centering
	\includegraphics[width=\linewidth]{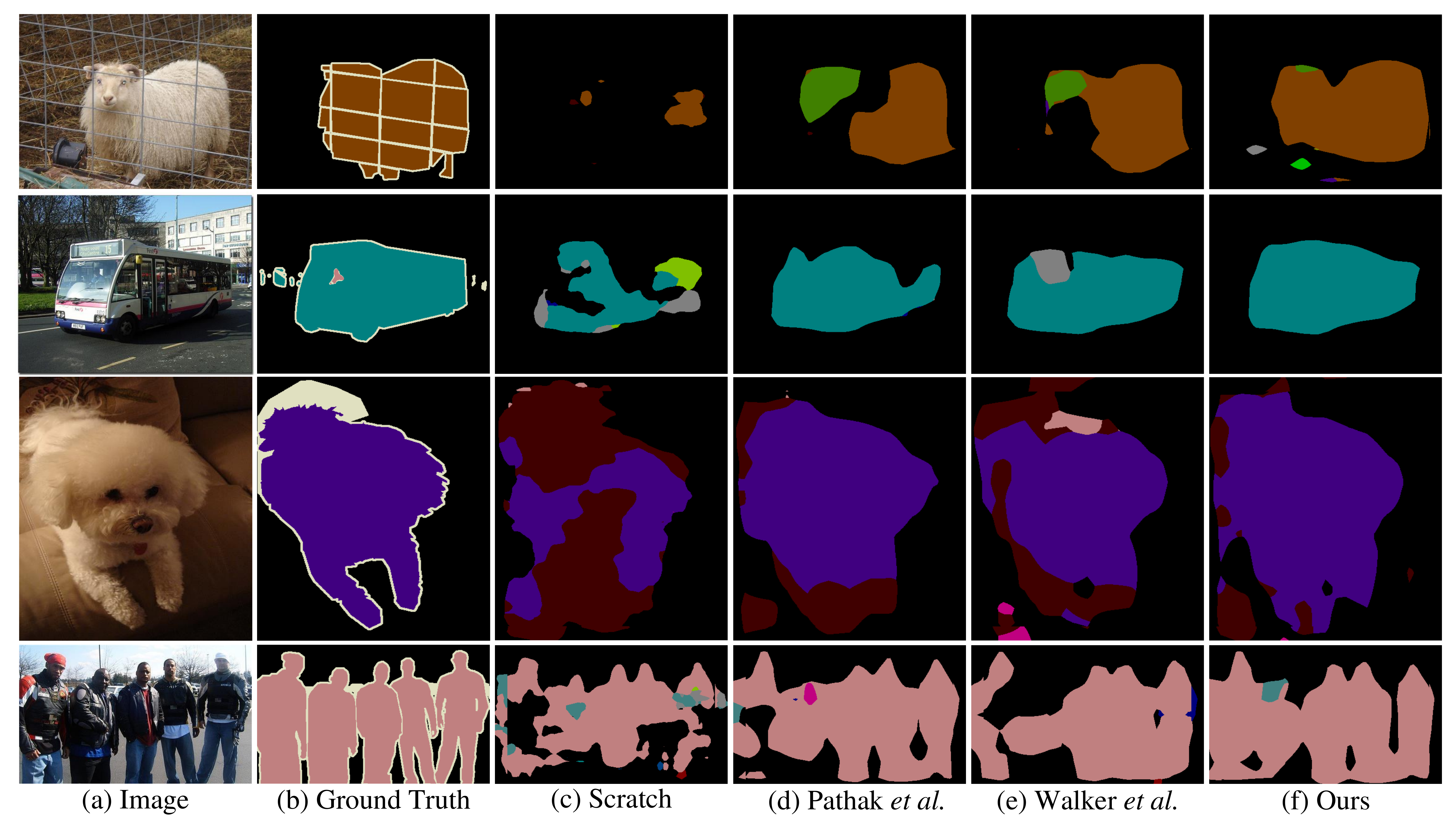}
	\caption{\label{voc} Visual improvements on the validation set of VOC2012 (AlexNet). }
\end{figure}

\begin{figure}[!t]
	\centering
	\includegraphics[width=\linewidth]{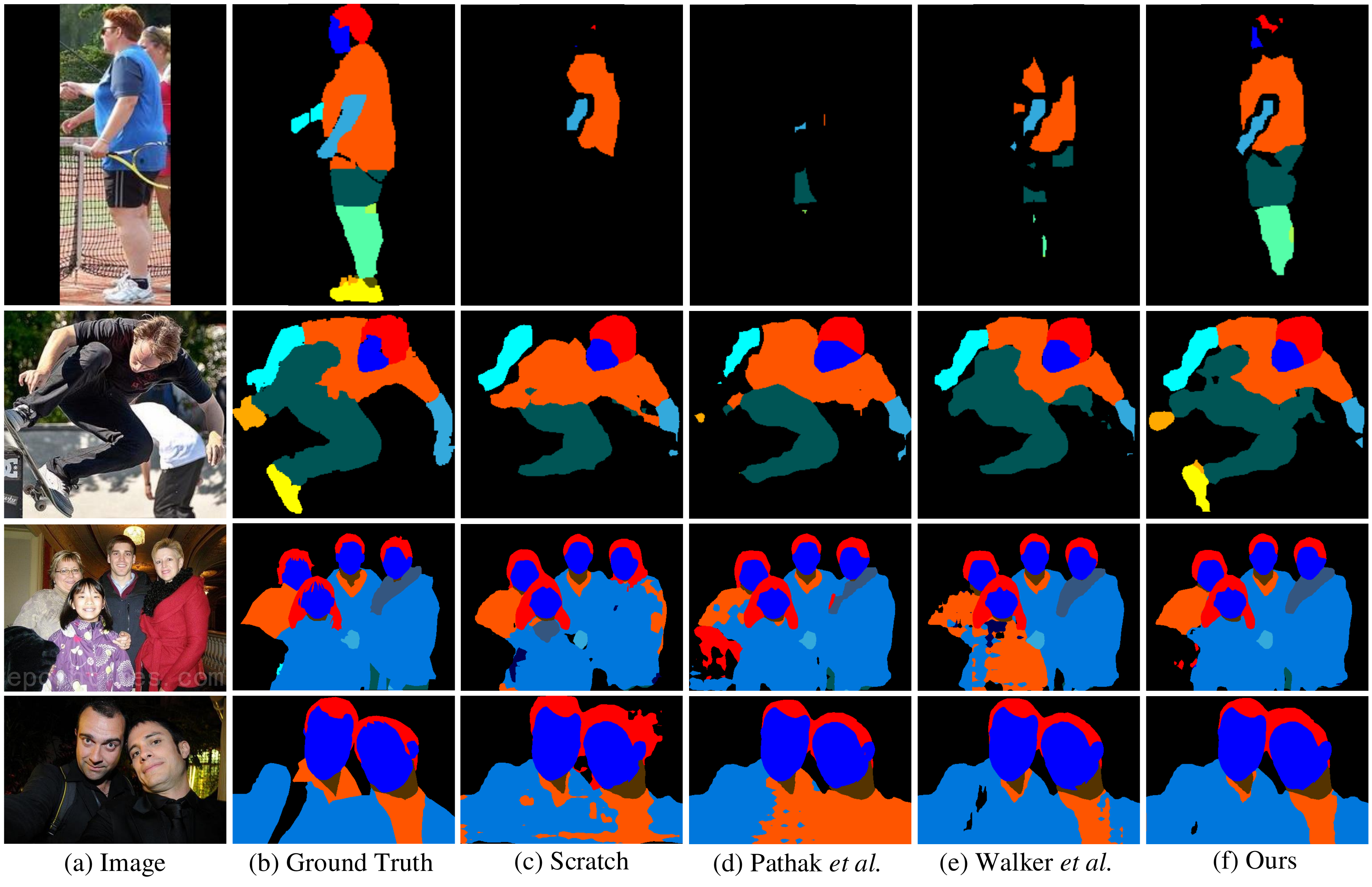}
	\caption{\label{lip} Visual improvements on the validation sets of LIP single-person and multi-person tasks (ResNet-50). }
\end{figure}

\noindent\textbf{Target Tasks.}
For the fine-tuning tasks on semantic segmentation and human parsing, we show the visual comparisons between our method and baselines in Figure~\ref{voc} and Figure~\ref{lip}, corresponding to PASCAL VOC 2012 and LIP datasets respectively.
When using our CMP pre-trained models, the fine-tuning results are more accurate and spatially coherent.
For example, as the first three rows of Fig.\ref{voc} show, baseline methods misclassify some parts of the sheep, bus, and dog, while our method produces spatially accurate and coherent results.
It is due to the kinematically-sound representations learned from CMP.

\end{document}